\documentclass[10pt,twocolumn,final]{article}

\usepackage{3dv}
\usepackage{times}
\usepackage{epsfig}
\usepackage{graphicx}
\usepackage{amsmath,amsfonts,bm}
\usepackage{amssymb}
\usepackage[round]{natbib}
\usepackage{latexsym}
\usepackage{multirow}
\usepackage{makecell}

\usepackage{hyperref}
\hypersetup{breaklinks=true,bookmarks=true,hidelinks}

\usepackage{glossaries}
\glsdisablehyper 
\makeglossaries
\newacronym{PSI}{PSI}{Photonic Systems Integration}
\newacronym{ACFR}{ACFR}{Australian Centre for Field Robotics}
\newacronym{CRIS}{CRIS}{Centre for Robotics and Intelligent Systems}
\newacronym{ACRA}{ACRA}{the Australasian Conference on Robotics and Automation}
\newacronym{ACRV}{ACRV}{Australian Centre for Robotic Vision}
\newacronym{USyd}{USyd}{the University of Sydney}
\newacronym{UQ}{UQ}{the University of Queensland}
\newacronym{QUT}{QUT}{the Queensland University of Technology}
\newacronym{UCSD}{UCSD}{the University of California, San Diego}
\newacronym{ANU}{ANU}{Australia National University}
\newacronym{IMOS}{IMOS}{the Integrated Marine Observation System}
\newacronym{URI}{URI}{the University of Rhode Island}
\newacronym{WHOI}{WHOI}{Woods Hole Oceanographic Institution}

\newacronym{NSF}{NSF}{National Science Foundation}
\newacronym{LIEF}{LIEF}{Linkage Infrastructure, Equipment and Facilities}

\newacronym{ICCP}{ICCP}{the International Conference on Computational Photography}
\newacronym{CVPR}{CVPR}{Computer Vision and Pattern Recognition}
\newacronym{TIP}{TIP}{Transactions on Image Processing}
\newacronym{TSP}{TSP}{Transactions on Signal Processing}
\newacronym{JFR}{JFR}{the Journal of Field Robotics}
\newacronym{ISCAS}{ISCAS}{International Symposium on Circuits and Systems}
\newacronym{TOG}{TOG}{Transactions on Graphics}
\newacronym{ICRA}{ICRA}{International Conference on Robotics and Automation}
\newacronym{IROS}{IROS}{Intelligent Robots and Systems}
\newacronym{RA-L}{RA-L}{Robotics and Automation Letters}

\newacronym{AUV}{AUV}{autonomous underwater vehicle}
\newacronym{UAV}{UAV}{unmanned aerial vehicle}
\newacronym{USV}{USV}{unmanned surface vehicle}
\newacronym{UGV}{UGV}{unmanned ground vehicle}
\newacronym{GPS}{GPS}{global positioning system}
\newacronym{SLAM}{SLAM}{simultaneous localisation and mapping}
\newacronym{SfM}{SfM}{structure from motion}
\newacronym{AR}{AR}{augmented reality}
\newacronym{VR}{VR}{virtual reality}
\newacronym{MR}{MR}{mixed reality}
\newacronym{CNN}{CNN}{convolutional neural network}
\newacronym{DNN}{DNN}{deep neural network}
\newacronym{IMU}{IMU}{inertial measurement unit}
\newacronym{TOF}{TOF}{time of flight}

\newacronym{MDSP}{MDSP}{multi-dimensional signal processing}
\newacronym{ROS}{ROS}{region of support}
\newacronym{DOF}{DOF}{degree-of-freedom}
\newacronym{RMS}{RMS}{root mean square}
\newacronym{RMSE}{RMSE}{root mean squared error}
\newacronym{SNR}{SNR}{signal-to-noise ratio}
\newacronym{CNR}{CNR}{contrast-to-noise ratio}
\newacronym{PCA}{PCA}{principal component analysis}
\newacronym{MSE}{MSE}{mean squared error}

\newacronym{FIR}{FIR}{finite impulse response}
\newacronym{IIR}{IIR}{infinite impulse response}
\newacronym{DFT}{DFT}{discrete Fourier transform}
\newacronym{FFT}{FFT}{fast Fourier transform}
\newacronym{PSNR}{PSNR}{peak signal-to-noise ratio}
\newacronym{FPGA}{FPGA}{field programmable gate array}
\newacronym{GPU}{GPU}{graphics processing unit}
\newacronym{ASIC}{ASIC}{application-specific integrated circuit}
\newacronym{BW}{BW}{bandwidth}

\newacronym{PSF}{PSF}{point spread function}
\newacronym{SPAD}{SPAD}{single-photon avalanche diode}
\newacronym{FOV}{FOV}{field of view}
\newacronym{BRDF}{BRDF}{bidirectional reflectance distribution function}
\newacronym{FWHM}{FWHM}{full width at half maximum}
\newacronym{LF}{LF}{light field}
\newacronym{2pp}{2pp}{two-plane parameterization}
\newacronym{MLA}{MLA}{microlens array}

\newacronym{RANSAC}{RANSAC}{random sampling and consensus}
\newacronym{DoG}{DoG}{difference of Gaussian}
\newacronym{SIFT}{SIFT}{scale invariant feature transform}

\newacronym{NIR}{NIR}{near-infrared}  
\newacronym{HOG}{HOG}{hisogram of oriented gradient}

\newacronym{SVM}{SVM}{support vector machine}
\newacronym{BoW}{BoW}{bag of words}

\threedvfinalcopy 

\def\threedvPaperID{80} 

\usepackage[utf8]{inputenc} 
\usepackage[T1]{fontenc}    
\usepackage{url}            
\usepackage{booktabs}       
\usepackage{amsfonts}       
\usepackage{nicefrac}       
\usepackage{microtype}      

\usepackage{enumitem,amssymb}
\usepackage{soul} 
\usepackage{subcaption}
\usepackage{wrapfig}
\DeclareMathOperator{\diag}{diag}
\DeclareMathOperator{\tr}{tr}
\DeclareMathOperator{\MSE}{MSE}

\begin{document}
\title{ Multiplexed Illumination for Classifying Visually Similar Objects}

\author{%
  Taihua Wang\\
{\tt\small twan8752@uni.sydney.edu.au}\\
{\small Dept of Aerospace, Mechanical and Mechatronic Engineering, University of Sydney, NSW 2006, Australia}
\and
  Donald G. Dansereau\\
{\tt\small donald.dansereau@sydney.edu.au}\\
{\small Sydney Institute for Robotics and Intelligent Systems, University of Sydney, NSW 2006, Australia}
}

\maketitle

\begin{abstract}
Distinguishing visually similar objects like forged/authentic bills and healthy/unhealthy plants is beyond the capabilities of even the most sophisticated classifiers. We propose the use of multiplexed illumination to extend the range of objects that can be successfully classified. We construct a compact RGB-IR light stage that images samples under different combinations of illuminant position and colour.  We then develop a methodology for selecting illumination patterns and training a classifier using the resulting imagery.  We use the light stage to model and synthetically relight training samples, and propose a greedy pattern selection scheme that exploits this ability to train in simulation. We then apply the trained patterns to carry out fast classification of new objects. We demonstrate the approach on visually similar artificial and real fruit samples, showing a marked improvement compared with fixed-illuminant approaches as well as a more conventional code selection scheme.  This work allows fast classification of previously indistinguishable objects, with potential applications in forgery detection, quality control in agriculture and manufacturing, and skin lesion classification.
\end{abstract}

\section{Introduction}
\label{section_introduction}

While image classification is a well explored area, the classification of visually similar objects remains challenging.  Forged and authentic paintings and currency, diseased and healthy plants, and benign and malignant skin lesions can all be so visually similar as to evade classification with even the most sophisticated modern techniques.  In this work we propose the use of a multiplexed active illumination source to facilitate the classification of visually similar objects.

The idea of extracting information from images by controlling illumination conditions is not new.  Prior works controlling the incident light spectrum have distinguished everything from types of alcoholic beverage to forged paintings \citep{asano2018coded, moran2016roles}. Varying the positions of illuminants is also well established: light stage technology spatially distributes and multiplexes light sources to measure visual appearance under a variety of conditions, delivering photo-real relightable models~\cite{guo2019relightables}. 

In this work, we combine the ideas of spectral and spatial multiplexing to improve classification of visually similar objects. While prior work has combined spectral and spatial multiplexing to classify materials \citep{kampouris2018icl, asano2018coded,wang2017joint}, our work differs in that it is concerned with object, not material classification. This allows us to distinguish complex non-homogeneous specimens like bills, paintings, lesions, and fruit, opening a new range of applications where visual classification previously failed.

\begin{figure}
\centering
   \begin{minipage}{0.2\textwidth}
     \centering
     \includegraphics[width=\linewidth,height=\linewidth]{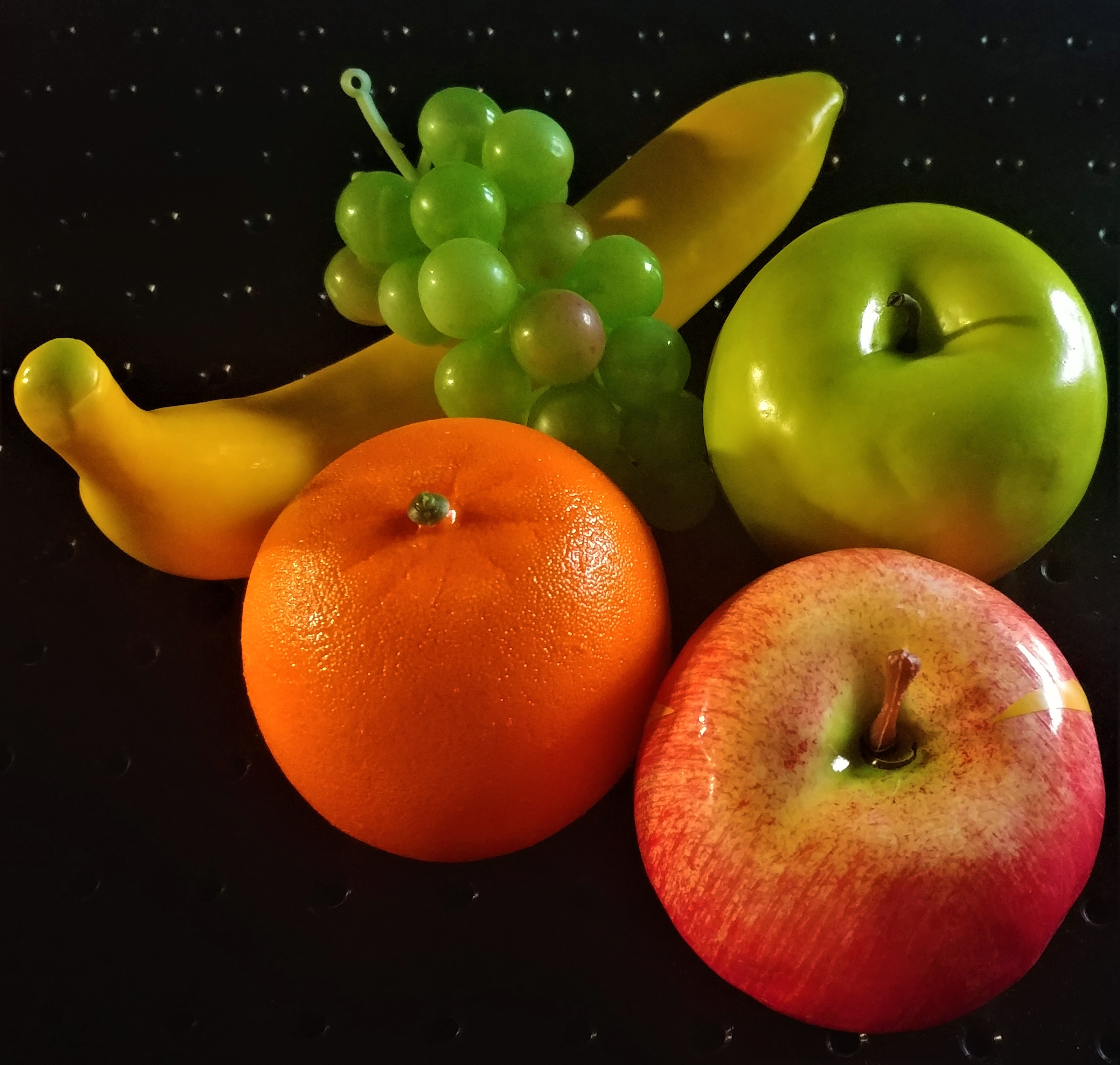}
   \end{minipage}
   \begin{minipage}{0.2\textwidth}
     \centering
     \includegraphics[width=\linewidth,height=\linewidth]{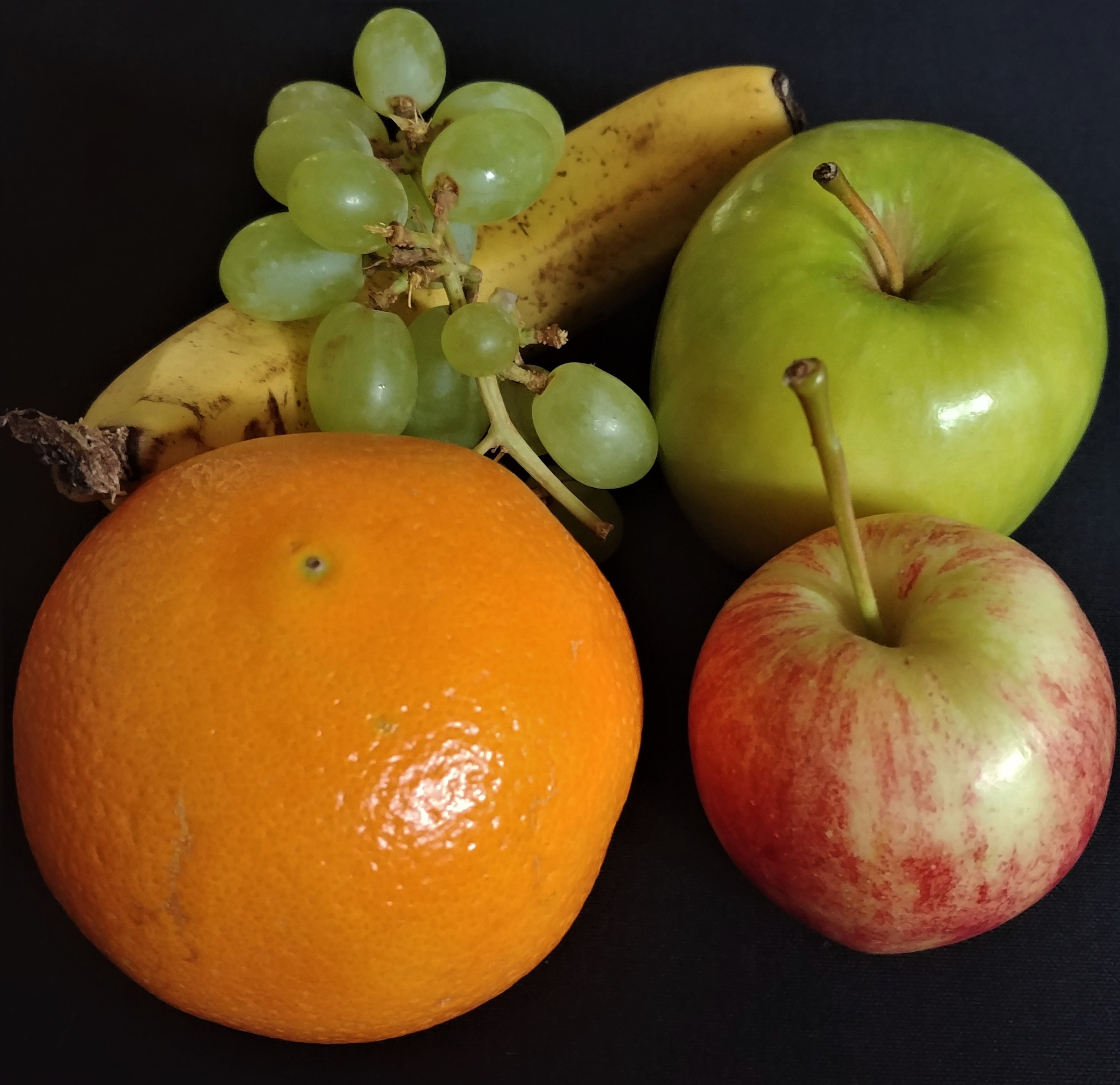}
   \end{minipage}
    \caption{Visually similar objects, like real and synthetic fruit, fool even the most sophisticated image classifiers. We develop an active multiplexed illumination scheme that allows conventional classifiers to distinguish these challenging examples, and propose a method for selecting illumination patterns that yields fast and accurate classification results.}
    \label{fig_fruit_examples}
\end{figure}

\begin{figure*}
\centering
   \begin{minipage}{0.65\textwidth}
     \subcaptionbox{Training}
     {\includegraphics[height=4.2cm]{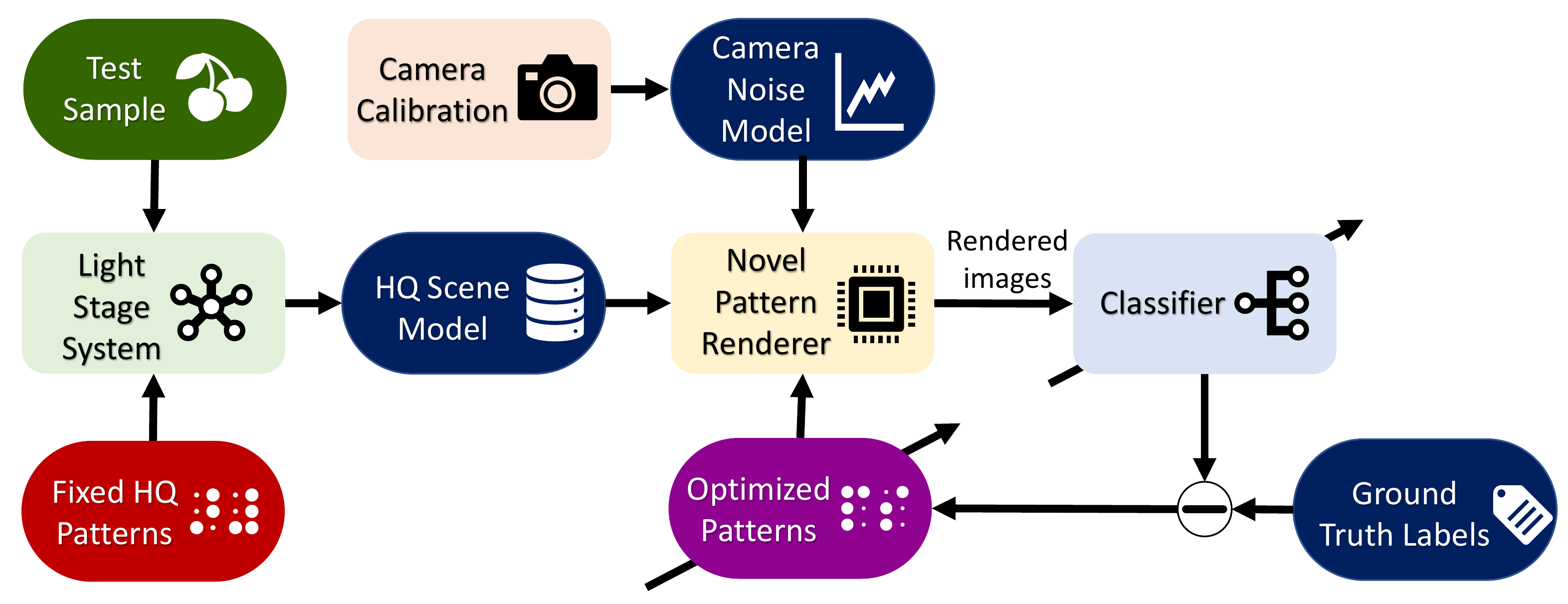}}
   \end{minipage}
      \begin{minipage}{0.29\textwidth}
      \subcaptionbox{Inference}
     {\includegraphics[height = 4.3cm]{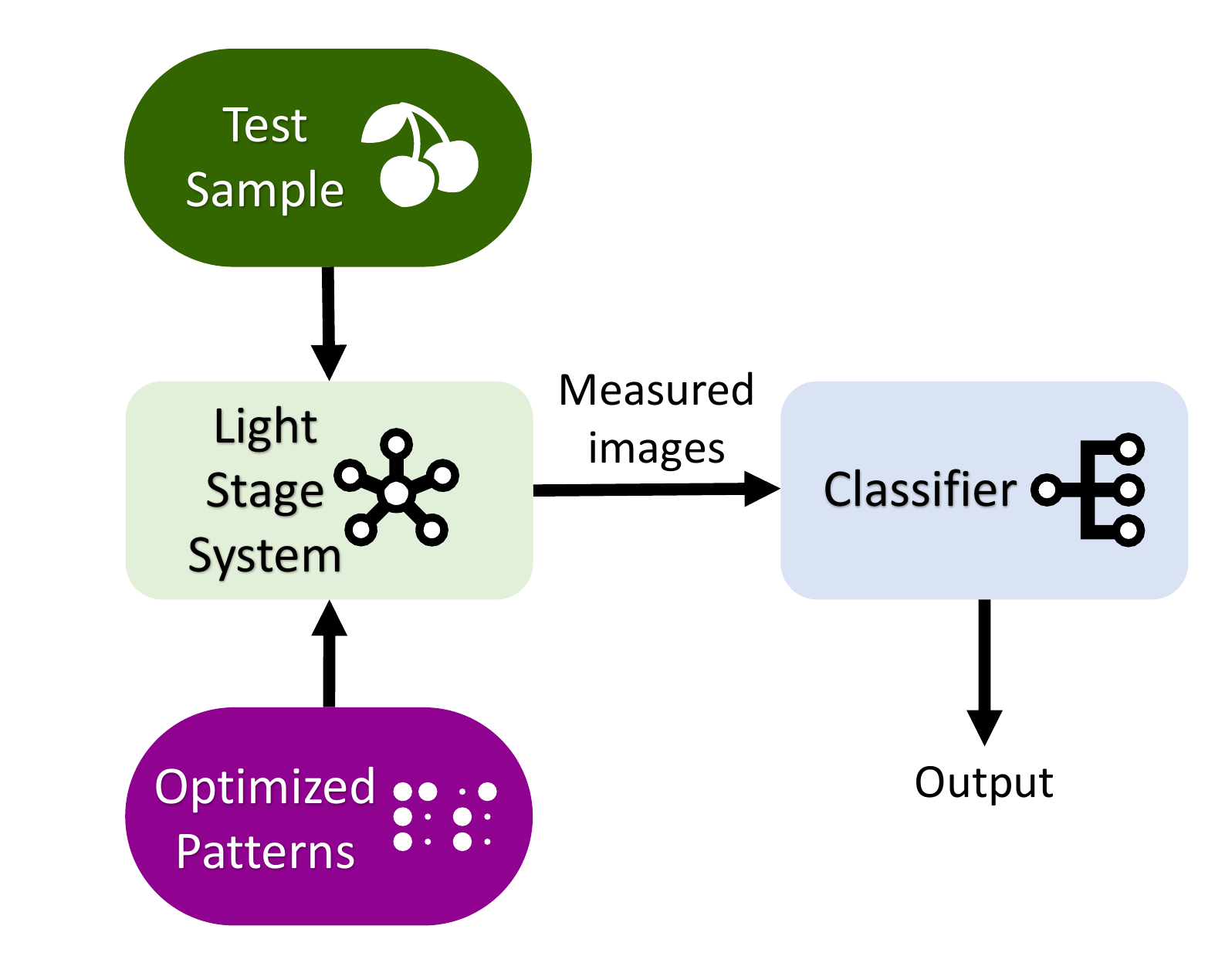}}
   \end{minipage}
  \caption{(a) We form high-quality models of training objects using the compact light stage, and use these in a physically accurate rendering pipeline to simulate different illumination conditions. This allows us to jointly train a classifier and optimize illumination patterns in simulation; (b) At inference time only the optimized illumination patterns are used, yielding fast and accurate classification.}
  \label{fig_system_diagram}
\end{figure*}

To this end, we construct a compact light stage featuring red, green, blue and \gls{NIR} colour channels, each repeated over eight illumination sites.  As in other multiplexed illumination schemes, the problem of selecting illumination patterns arises.  \Gls{SNR}-optimal codes are often assumed to maximize system-level performance in such scenarios \citep{mitra2014framework}. However, in this work we show that multiplexing codes can be selected to deliver greater classification accuracy and speed than are possible with \gls{SNR}-optimal codes.  

For code selection, we propose a methodology that uses the light stage in two ways: first we model and synthetically relight training samples to allow joint pattern selection and classifier training in simulation. Then we use the trained patterns and classifier to quickly classify previously unseen samples. 

Our key contributions are:
\begin{itemize}[noitemsep,topsep=0pt]
\item We propose a multiplexed illumination scheme that allows visually similar objects to be distinguished by conventional classifiers,
\item We use light stage capture and noise synthesis to simulate realistic images under different lighting patterns, allowing the training of multiplexing patterns and classifiers in simulation, 
\item We propose a time-efficient greedy optimization scheme for selecting illumination patterns, and
\item We demonstrate our method outperforming na\"ive fixed-illumination and \gls{SNR}-optimal multiplexing codes in terms of classification accuracy and speed.
\end{itemize}

We are also releasing a dataset containing 16000 images of real and synthetic examples of each type of fruit depicted in Fig.~\ref{fig_fruit_examples}. Images were collected under a variety of illumination patterns including single and multiple-illuminant states. Because it allows the evaluation of novel multiplexing schemes, we believe the dataset and methodology described here might form the basis for extensive follow-on work. Code and dataset are available at \url{https://roboticimaging.org/Projects/LSClassifier/}.

The remainder of the paper is structured as follows: Sect.~\ref{sect_related_work} highlight related work and the knowledge gap that this paper fills; Sect.~\ref{sect_method} lays out the details of the proposed approach; Sect.~\ref{sect_results} provides system validation and experimental results; and Sect.~\ref{sect_conclusions} draws conclusions and indicates directions for future work.

\section{Related work}
\label{sect_related_work}

\begin{figure*}
\centering
    \begin{minipage}{0.3\textwidth}
     \centering
     \subcaptionbox{}
     {\includegraphics[width=0.93\linewidth,height=6cm]{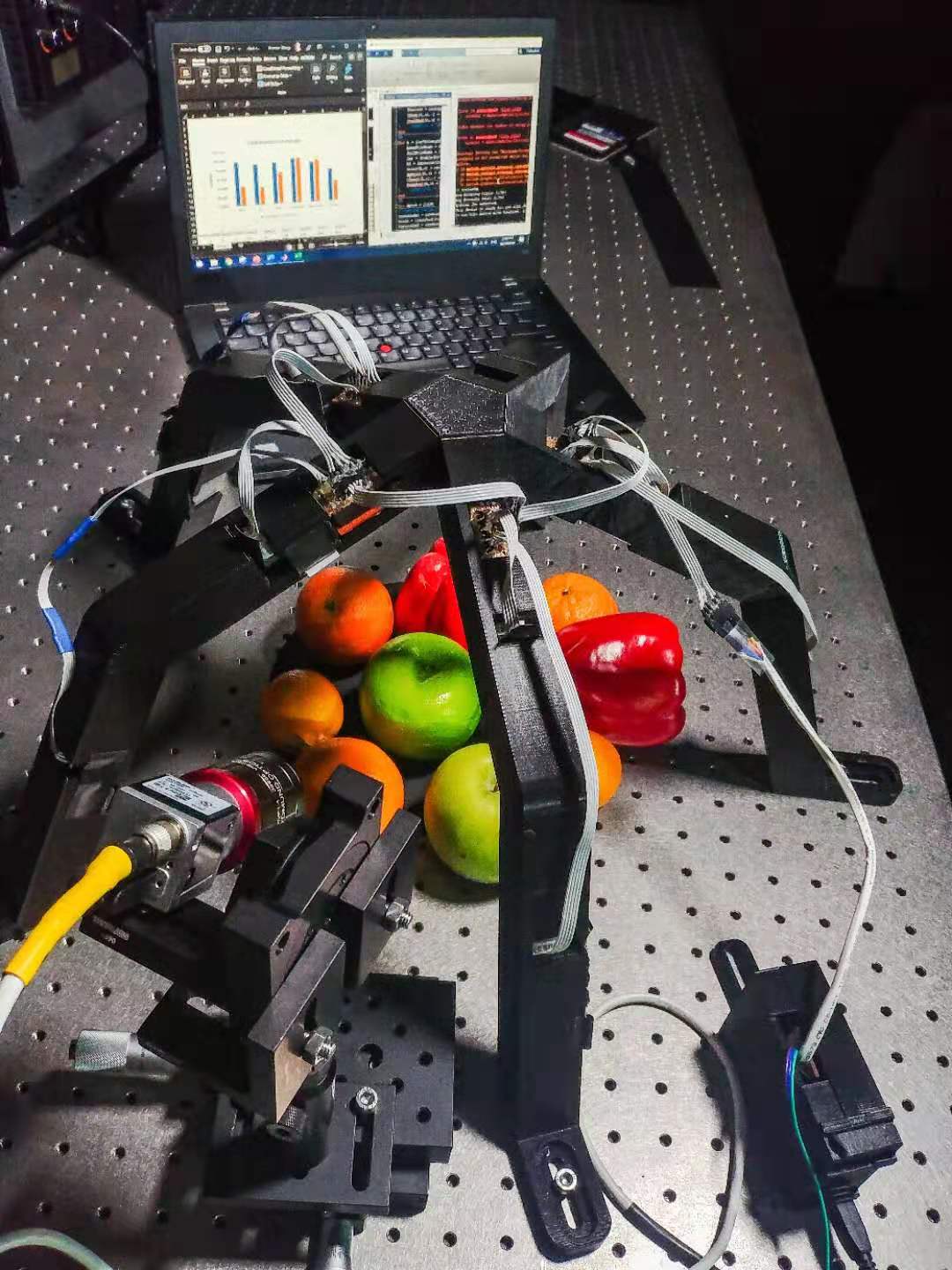}}
   \end{minipage} \hspace{1.5em}  
   \begin{minipage}{0.3\textwidth}
     \centering
     \subcaptionbox{}
     {\includegraphics[width=0.93\linewidth,height=6cm]{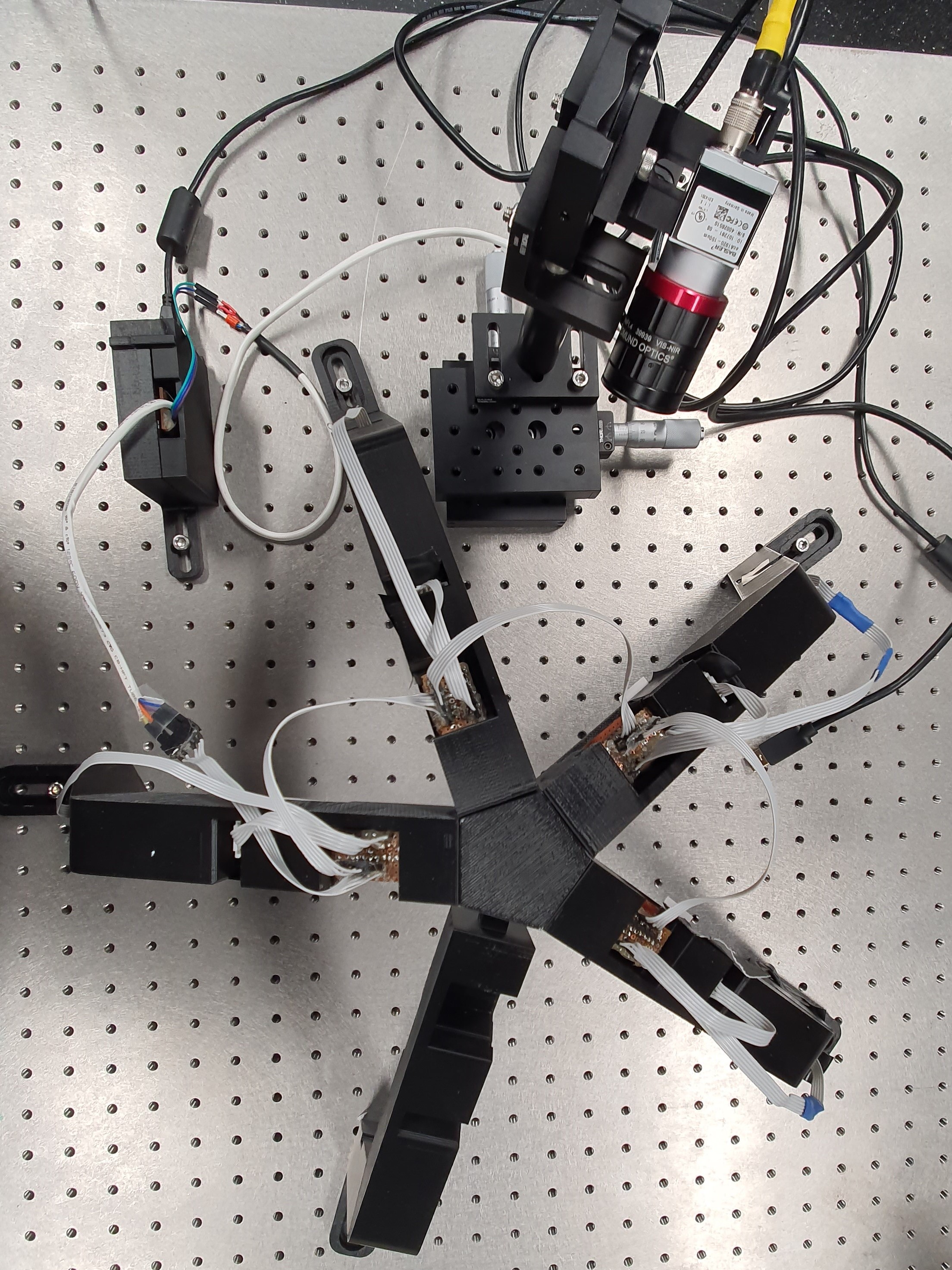}}
   \end{minipage} \hspace{-1.2em}
   \begin{minipage}{0.3\textwidth}
     \centering
     \subcaptionbox{}
     {\includegraphics[height=6cm]{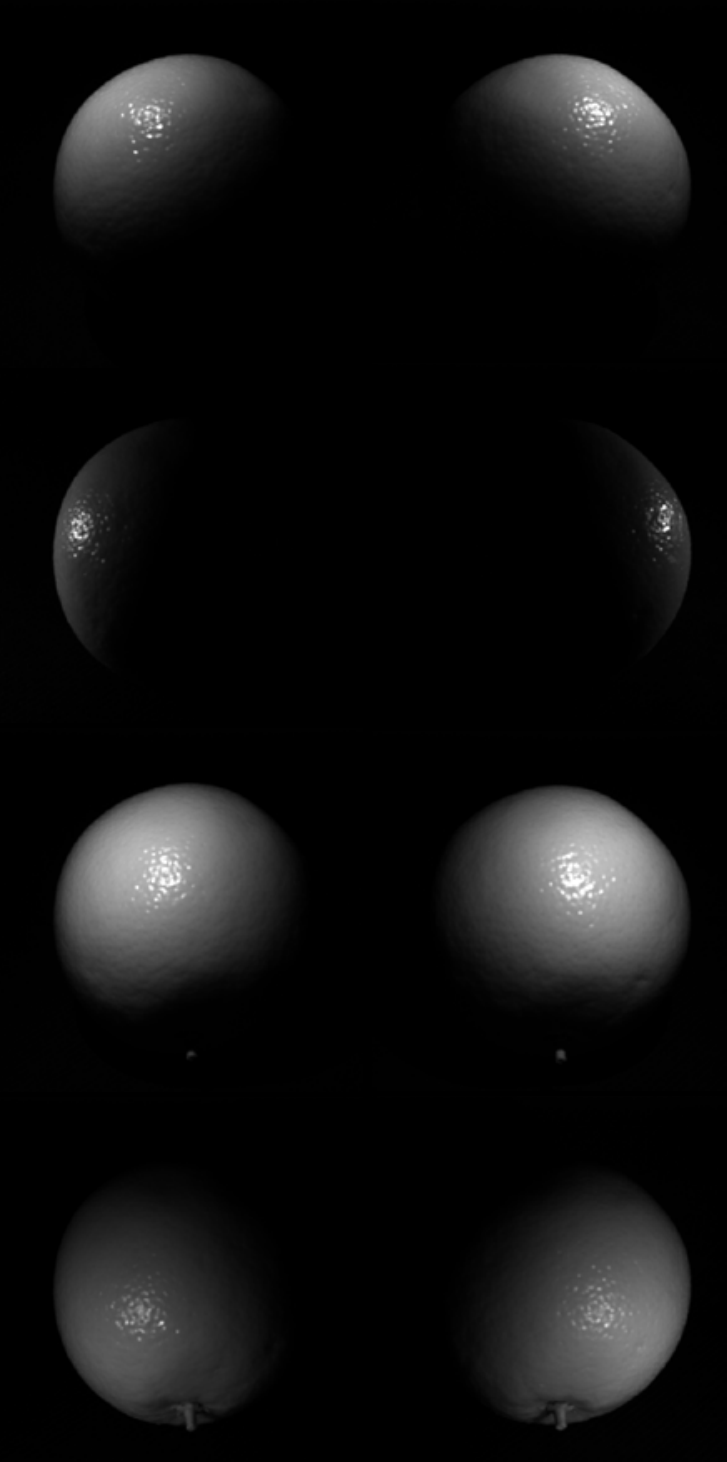}}
   \end{minipage}
    \caption{The compact light stage shown (a) in profile and (b) from above; (c) depicts typical images obtained with a single illumination site active at a time.}
    \label{fig_light_stage}
\end{figure*}

This work combines the concepts of coded (multiplexed) illumination, multispectral illumination, and image relighting to improve image classification.  While prior work has established the utility of these techniques for applications in computer graphics, BRDF acquisition, and material classification, none to our knowledge has yet combined them for improving classification of visually similar objects.

Light multiplexing for enhanced image quality \citep{schechner2007multiplexing} has shown that $N$ single-illuminant images can be obtained at higher \gls{SNR} by capturing $N$ coded illumination states, followed by demultiplexing. The work establishes multiplexing codes that yield optimal \gls{SNR}, with follow-on works refining treatment of effects like Poisson noise and saturation \citep{ratner2007illumination}. These works underpin much of our understanding of computational imaging \citep{mitra2014framework}, which can be seen as delivering improved image quality through multiplexed light capture. A key contribution of our work is to apply multiplexing to improve image classification. Like other works considering applications beyond intensity estimation \citep{alterman2010multiplexed}, we show that \gls{SNR}-optimal codes do not generally yield the best overall system performance.

Light stage capture and relighting \citep{sato2005using, wenger2005performance, debevec2012light} employs multiplexed illumination across a diversity of illuminant poses to capture high-fidelity relightable visual models. Modern examples can produce photo-real relit images of very complex subjects including human faces \citep{guo2019relightables}. In this work we adopt the light stage in two ways: we build a relightable model of the training samples and use this capability to select multiplexed illumination patterns for image classification; then we apply the resulting patterns using the light stage for fast inference on previously unseen samples.

Applying spectral control to improve material classification is also well established \citep{zhaoxiang2007apple, blasinski2017designing}, with spectra featuring \gls{NIR} components proving especially useful \citep{nagata2006bruise,roy2016bruise, teena2014near, lu2011vis,nicolai2006non,wachs2010low}. Similarly, multiplexed multispectral illumination has generally dealt with rendering \citep{legendre2016practical} or material classification \citep{kampouris2018icl,asano2018coded}. The later employs coded illumination to carry out single-image pixel-level material classification, and \cite{wang2017joint} use a light stage for discriminating materials using a few images. Our focus is on classifying objects rather than materials, opening new domains involving heterogeneous, complex objects.

This is the first work to our knowledge that combines the concepts of multispectral and multiplexed illumination to enable object-level classification of very challenging samples. We develop a time-efficient means of deriving multiplexing codes that leverages light stage relighting to carry out the process in simulation. We then employ the light stage at inference time to quickly capture multiple images that allow us to accurately classify objects that would otherwise be visually indistinguishable. This allows classification of objects with heterogeneous materials, opening the technique to new applications requiring object-level discrimination. 

\section{Modelling, Training, and Inference}
\label{sect_method}

The system proposed in this paper consist of two stages as depicted in Fig.~\ref{fig_system_diagram}. In training, we build high-quality relightable models of our test samples, using the custom-built light stage depicted in Fig.~\ref{fig_light_stage}. We use these models to select fast and informative multiplexing patterns and simultaneously train a classifier. At inference time, test samples are measured using only the optimized patterns, yielding fast, accurate classification results.

\subsection{Scene Model Acquisition}
\label{sect_model_acquisition}

We begin by collecting relightable models of a variety of samples, each under a variety of poses. For this we turn the light stage to the task it was originally designed for: multiplexing illumination to build relightable models \citep{debevec2012light}.  Because we want to synthesize, in simulation, noise representative of inference-time conditions, it is important for these models to be captured at a high \gls{SNR}. We thus operate the camera using long exposures, and take multiple exposures of each sample / pose which we average together to further improve \gls{SNR}. Then the synthetic noise added in simulation will dominate over the small amount of noise present in the high-quality scene models.

The model acquisition process is much more time-consuming than inference-time image capture, but is worthwhile because it enables fast simulation of a vast variety of illumination states. The total time required to capture high-quality scene models and then simulate thousands of fast illumination patterns is much less than the time it would take to directly test those patterns. 

\subsection{Camera Noise Characterization}
\label{sect_camera_cal}

To render physically accurate relit images we require a model of the camera's noise parameters, allowing us to synthetically introduce realistic noise into the rendered images.  We characterize the camera when it is configured for inference, including the optics, aperture, and focus, as well as exposure time and gain settings.  We then generalize the resulting calibrated camera noise model to other camera settings, allowing us to evaluate, in simulation, a variety of additional imaging scenarios.

Because in this application we are evaluating illumination patterns and test samples that yield a variety of image intensities, it is especially important that our noise model account for intensity-varying Poisson-distributed photon noise. We thus adopt the affine noise model \citep{schechner2007multiplexing}, that describes the total noise at a pixel as 
\begin{equation}
    \sigma^2 = \sigma^2_p\bar{I}  + \sigma^2_r,
    \label{eq_affine_noise}
\end{equation}
where $\sigma^2_r$ is the variance of the fixed read noise, $\sigma^2_p$ is the scaling constant for the variance of the intensity-dependent photon noise, and $\bar{I}$ is the mean intensity of each pixel.

\begin{figure}
\centering
   \begin{minipage}{0.33\textwidth}
     \centering
     \subcaptionbox{Characterizing camera noise}
     {\includegraphics[width=\linewidth,page=1]{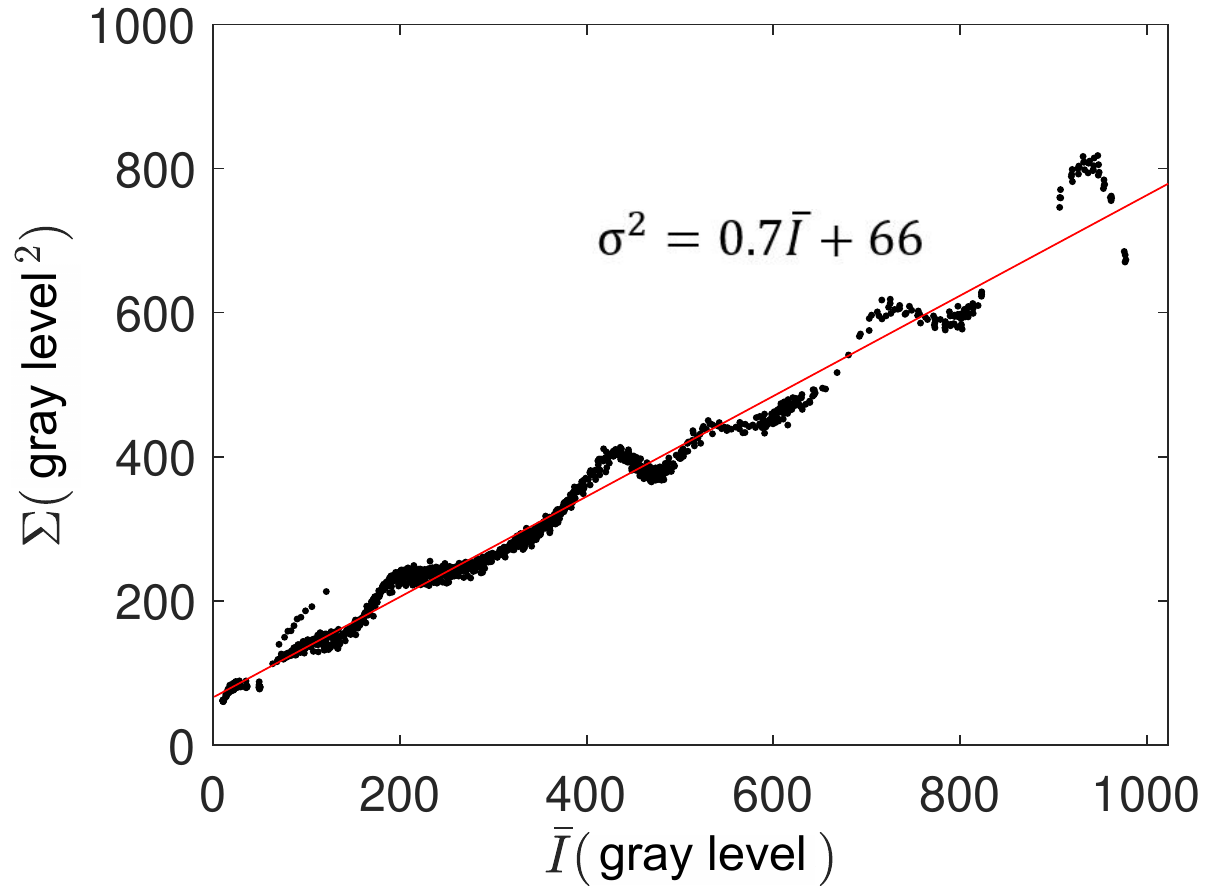}}
   \end{minipage}
      \begin{minipage}{0.33\textwidth}
     \centering
     \subcaptionbox{Verifying noise model and synthesis}
     {\includegraphics[width=\linewidth,page=2]{camnoisecurve.pdf}}
   \end{minipage}
    \caption{characterizing imaging noise: the Basler acA1920-150um at 15~dB gain and 30~ms exposure time yields a characteristic (a) that is well described by an affine noise model~\eqref{eq_affine_noise}; (b) We generalize this model following~\eqref{eq_generalized_noisevar} to three noise characteristics $\Sigma_{1..3}$ corresponding to typical camera gain and exposure settings~\eqref{sigma_levels1}..\eqref{sigma_levels3}. The measured (dark) points conform closely to the predicted noise curves (solid lines). We validate noise synthesis by rendering scenes for each camera setting, yielding the lighter points that agree well with the models.}
    \label{fig_noisecurve}
\end{figure}

There are well established methods for calibrating camera noise \citep{liu2006noise,schechner2007multiplexing,ratner2007illumination,alterman2010multiplexed}. Because we have a controlled illumination source in the form of a compact light stage, we calibrate our model by capturing images of a test pattern with smoothly varying reflectivity over a variety of illumination levels. We then find pixel means and variances over multiple exposures, and fit the affine noise model to the resulting observations. An example characterization for the camera used in this work is shown in Fig.~\ref{fig_noisecurve}.

We generalize the noise model to different camera settings by considering the ratio of a new gain $G$ and exposure time $E$ to their calibrated levels $G_0, E_0$. The total variance for the generalized noise model becomes
\begin{equation}
    \Sigma = \frac{G^2}{G_0^2}(\frac{E}{E_0} \sigma_p^2 \bar{I} + \sigma_r^2).
    \label{eq_generalized_noisevar}
\end{equation}
Note that while photon noise varies with both gain and exposure time, read noise is well approximated as varying only with gain. 

\subsection{Simulating Novel Illumination Patterns}
\label{sect_rendering}

For a light stage with $N$ illumination sites, we describe an illumination state with a column vector $\bm{w}$ of length $N$, where $w_i \in [0,1]$.  Each relightable model contains as many images as there are illuminants, $N$. For images with a total of $N_{pix}$ pixels, we reshape each into a column vector and stack them columnwise into a model $L$ of size $N_{pix} \times N$.  Then, rendering a noise-free image can be written as 
\begin{equation}
    \bar{I} = L \mathbf{w},
\end{equation} 
where $\bar{I}$ indicates the mean image intensity at each pixel. 

To simulate noise at a prescribed camera gain and exposure, we use the generalized noise model~\eqref{eq_generalized_noisevar} to yield a per-pixel physically derived variance $\Sigma$. For each pixel we then draw from a Gaussian distribution 
\begin{equation}
    I \sim \mathcal{N}(\bar{I},\Sigma),
    \label{eq_gauss_noise}
\end{equation}
clipping the resulting value to lie within the gray level range of the camera.

\subsection{Joint Pattern Selection and Classifier Training}
\label{sect_training}

The ability to simulate illumination states with physically realistic noise allows us to train and evaluate classifiers under a variety of novel conditions. In particular, we can collect a sequence of $M$ images, each corresponding to a different illumination state $\bm{w}$. Taken together, these acquisition states form an $N \times M$ illumination matrix $W$. Gathering more images takes more time, but can increase classifier performance. The goal of the training process is thus to select the illumination matrix $W$ and classifier that together maximize classifier accuracy and speed.

Fig.~\ref{fig_system_diagram} shows how rendered images drive the joint optimization of the illumination matrix and the classifier. In the main training loop, a set of illumination patterns drives the rendering of a new set of images. These are used to train and evaluate a classifier, and this process repeats, optimizing towards improved classifier performance by selecting better illumination matrices.

\begin{figure}
\centering
    \includegraphics[width=0.8\linewidth]{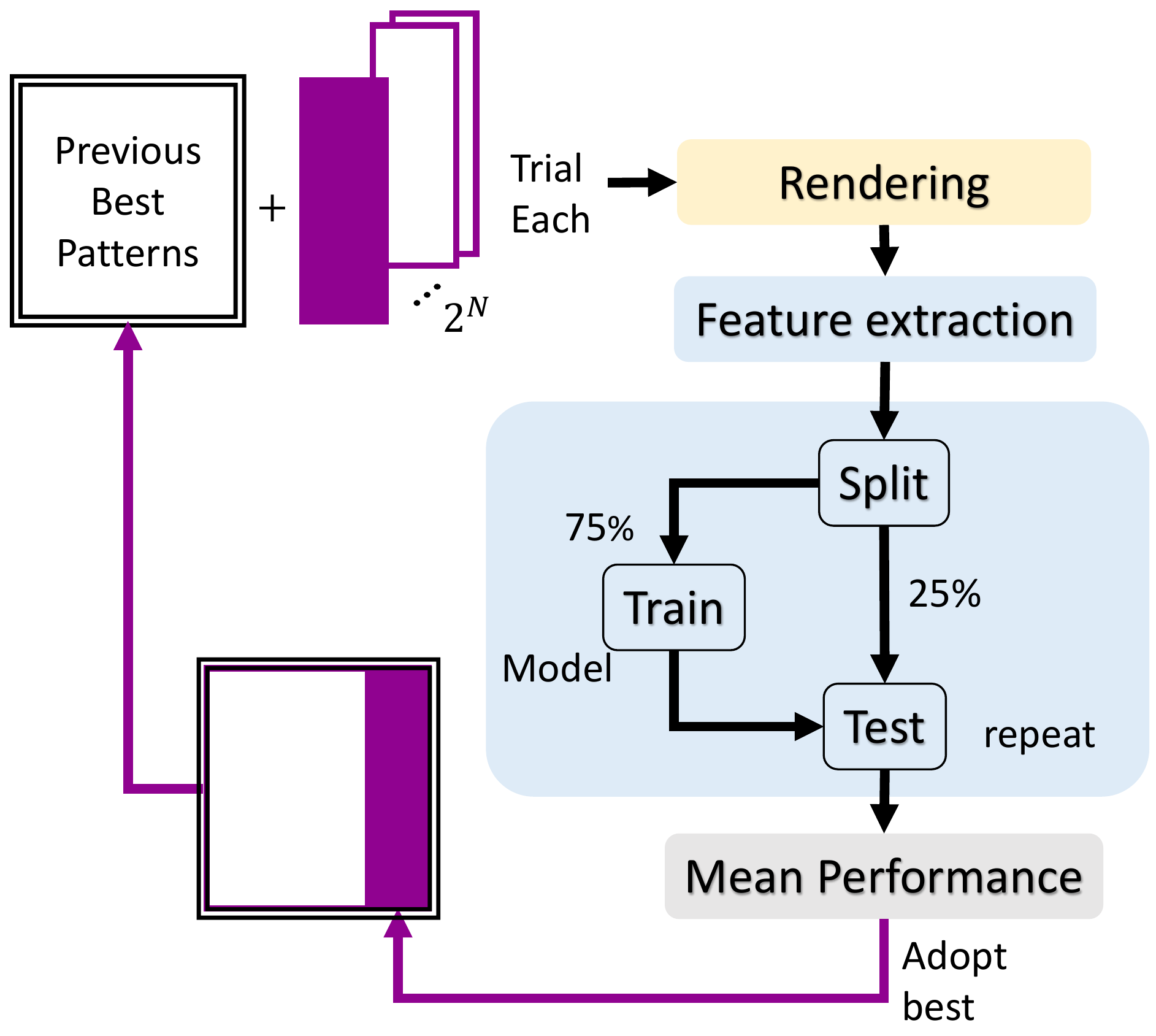}
    \caption{Greedy pattern selection: the illumination matrix (top left) is incrementally grown one pattern at a time. For each new column, we perform an exhaustive search over all $2^N$ possible new patterns, for each rendering the corresponding relit imagery, training and testing a classifier, and selecting the one that yields the highest overall accuracy. Train and test are repeated to make best use of the dataset.}
    \label{fig_pattern_selection}
\end{figure}

We anticipate a variety of approaches could be adopted for this optimization process. In this work we propose a greedy scheme, depicted in Fig.~\ref{fig_pattern_selection}, that incrementally grows a multiplexing matrix one column at a time. This begins by determining the best-performing single-image illumination pattern. It then builds on this by incrementally adding images, selecting the best-performing illumination pattern for each. The process ends when no illumination pattern improves performance of the system. The practitioner can then choose to use the whole illumination matrix, or to use a subset to obtain faster but less accurate results.

For our implementation, we evaluate individual patterns exhaustively. That is, for $N$ illumination sites, we test $2^N$ illumination states. For an $M$-pattern sequence, we test a total of $M 2^N$ illumination matrices. Each pattern is evaluated by rendering relit versions of all available training scenes, and splitting the result to train and test the classifier. To make maximal use of the available models, we repeat this process for different random selections of training and test data. We also re-render images in each pass, to avoid overfitting to a particular instance of randomly selected noise.

We employ a conventional multi-class \gls{SVM}-based classifier, though we expect the method will generalize well to other classifiers including \glspl{DNN}. To drive the \gls{SVM}, we concatenate descriptors computed on regular grids on each of the $M$ differently illuminated images.  We do not employ a \gls{BoW} as these cluster and equate features with similar appearance, an undesirable trait when classifying visually similar classes. Rather we rely on a diversity of sample poses to provide rotational and translational invariance, and the consistency of the light stage camera pose to limit apparent scale variations.

For a given illumination matrix $W$, different strategies can be used for controlling camera exposure and gain. We fix the camera exposure to a desired setting that will ultimately determine the acquisition and thus inference speed for the system.  For each matrix under consideration, we select a single camera gain appropriate for the brightest image in the set, yielding a single-setting acquisition process.

Inference is carried out on previously unseen samples using the optimized illumination matrix and classifier. As depicted in Fig.~\ref{fig_system_diagram}, the illumination patterns drive the light stage to quickly capture a set of images which are passed to the classifier.  The inference process can be much faster than the modelling process, because the illumination patterns are trained to operate with a fast camera exposure time, and are selected to yield accurate results with only a few images.

\begin{figure}
  \centering
  \includegraphics[width=\linewidth]{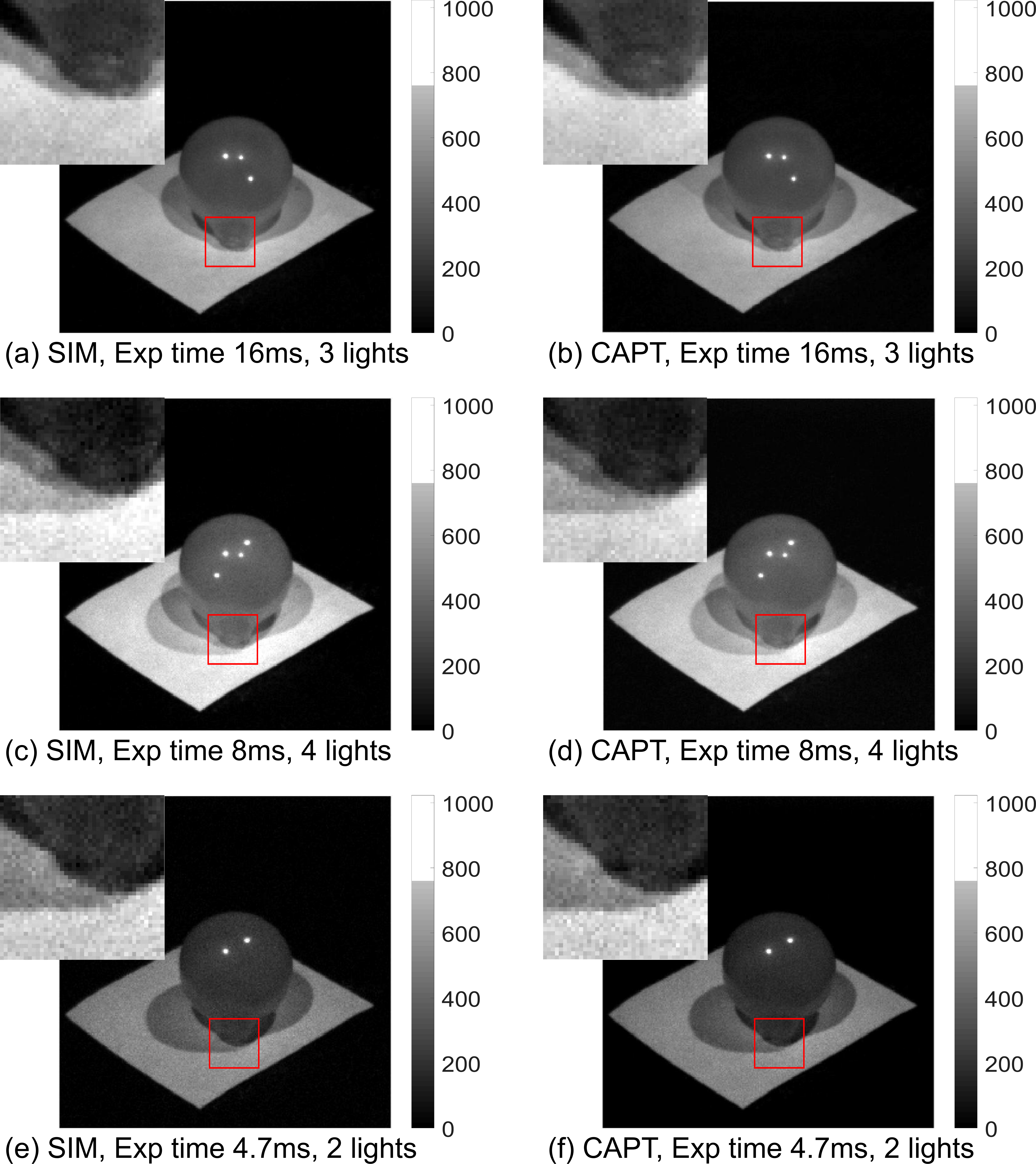}
   \caption{We validate rendering by generating imagery at different illumination conditions and camera settings (left), and comparing with the corresponding captured imagery (right).}
    \label{fig_noise_rendering}
\end{figure} 

\subsection{SNR-Optimal Illumination Patterns}
\label{sect_snr_optimal_pattern_selection}

As an alternative to the greedy pattern selection scheme described above, we also present a method based on the theory of multiplexed illumination, in which the goal is to estimate a set of single-illuminant images from a set of coded multiple-illuminant images \citep{schechner2007multiplexing}. Multiplexing illumination in this way boosts the \gls{SNR} of the recovered images, and \gls{SNR}-optimal multiplexing is generally held to yield maximal performance in computational imaging systems. However, as we will show, this method does not adapt illumination patterns to the problem space, yielding lower performance compared with our proposed greedy scheme.

To correctly account for a specific calibrated camera noise characteristic including Poisson noise, we adopt an optimization-based approach that adjusts the multiplexing matrix $W$ to maximize \gls{PSNR}, assuming an average scene reflectance $\bar{R}$ \citep{mitra2014framework}.  Starting with the identity matrix, we employ a stochastic optimization scheme in which each iteration perturbs the previous best matrix and evaluates the result. Checking the condition number allows us to reject non-invertible matrices. 

To evaluate a matrix $W$ we estimate the noise variance expected in the measured imagery by generalising~\eqref{eq_affine_noise} as
\begin{equation}
    \Sigma_W = \diag\left(\sigma^2_p\bar{R}\sum_{col}{W}+\sigma^2_r\right),
    \label{eq_snr_noisevar} 
\end{equation}
where the summation over columns of $W$ estimates the overall brightness for each illumination state, and the average scene reflectance $\bar{R}$ converts this to an average pixel intensity. We then estimate the \gls{MSE} of the recovered single-illuminant images following \cite{alterman2010multiplexed}
\begin{equation}
    \MSE = \frac{1}{n}\tr\left[({W}^T{\Sigma_W}^{-1}{W})^{-1}\right].
    \label{eq_mse_minimize}
\end{equation}
Driving the estimated \gls{MSE} to a minimum yields the \gls{PSNR}-optimal multiplexing matrix.

\section{Results}
\label{sect_results}

To evaluate our methodology we constructed a compact light stage prototype as depicted in Fig.~\ref{fig_light_stage}. The prototype features a five-leg design, with a 223~mm radius and 240~mm height. Its eight illumination sites are distributed across four of its legs, four mounted on upper leg segments, and four mounted on lower.  

Each illumination site has four LEDs centered on Red (615~nm), Green (515~nm), Blue (460~nm), and \gls{NIR} (850~nm) colour bands. A diffuser on each unit softens specular highlights and reduces apparent position variation with colour. LED control is via an extensible daisy-chained circuit.

We employ a Basler acA1920-150um monochrome machine vision camera with an Edmund Optics NIR-VIZ 6mm infrared-compatible lens. We crop the captured images to correspond to the working area, yielding square images 800 pixels on side. The camera and LEDs are hardware-triggered by a microcontroller to maintain synchronization.

We evaluate our system using the five types of synthetic and real fruit depicted in Fig.~\ref{fig_fruit_examples}. The two sets of experiments in this paper employ different sample sets, and evaluation on captured imagery uses samples not seen during training. A high quality relightable model of each training sample was collected over 20 different poses, as described in Sect.~\ref{sect_model_acquisition}.  We employed a single-illumination approach to model acquisition, though this could be accelerated using multiplexing as in~\cite{schechner2007multiplexing}. In total we collected 6400 images for training, and another 12800 for testing, 6400 for each of the proposed and \gls{SNR}-optimal approaches.

\subsection{Validating Rendering and Noise Synthesis}

Here we validate the scene model acquisition, camera noise characterization, and physically realistic rendering processes described in Sects.~\ref{sect_model_acquisition}, \ref{sect_camera_cal}, and \ref{sect_rendering},   

Camera calibration proceeds by imaging a monochromatic object under increasing illumination intensities.  We characterize the camera at 15~dB gain and 30ms exposure time, collecting 120 images at each of a range of illumination intensities, yielding the characteristic shown at the top of Fig.~\ref{fig_noisecurve}. Following~\eqref{eq_affine_noise}, a line fit yields the affine noise model
\begin{equation}
    \sigma^2=0.7\Bar{I}+66.
    \label{15dbnoise}
\end{equation}

We wish to generalize the camera's characterization to other gain and exposure settings. We thus select three typical exposure times of 84, 42, and 22.5~ms, and pair each with a corresponding gain setting of 6, 12, and 17.5~dB, with the latter approaching the camera's maximum gain.  We label each of these three settings $\Sigma_{1..3}$, and use \eqref{eq_generalized_noisevar} to estimate the camera's noise characteristic at each:
\begin{align}
\Sigma_1&: \sigma_p^2 = 0.25, \, \sigma_r^2 = 8.35, \label{sigma_levels1}\\
\Sigma_2&: \sigma_p^2 = 0.50, \, \sigma_r^2 = 33.23, \mathrm{and}\\
\Sigma_3&: \sigma_p^2 = 0.94, \, \sigma_r^2 = 117.37. \label{sigma_levels3}
\end{align}

We validate each of these new noise models by characterizing the camera at each setting $\Sigma_{1..3}$. The results, seen in the bottom of Fig.~\ref{fig_noisecurve}, show good agreement between each model (solid line) and measured characteristic (dark points). A slight difference in the read noise is evident with the noisier setting $\Sigma_3$, but the difference is within a few gray levels and the overall fit is good. 

We also validate the noise introduced in the rendering pipeline by simulating images at each of the camera settings $\Sigma_{1..3}$. Repeating each rendering and taking statistics over the set allows us to plot noise characteristics for the rendered images, seen as faint points in Fig.~\ref{fig_noisecurve}, and showing good agreement with the targeted characteristic in each case. Note the decrease in variance near the top end of the curves is due to saturation.

To further validate the rendering process, we synthesized and captured images across several illumination states and camera settings.  Typical examples are shown in Fig.~\ref{fig_noise_rendering}, showing good agreement between rendered and captured images. As exposure time decreases, top to bottom, we perceive a clear decrease in \gls{SNR}, as expected.

\subsection{Pattern and Classifier Training}

Our light stage has eight illumination sites, each having four LEDs.  We could optimize patterns over all combinations of the 32 LEDs.  Although we expect this would yield strong results, it also represents a massive search space, even for the efficient greedy method proposed here.  We thus proceed by optimizing patterns over the eight illumination sites, repeating the spatial pattern across the four colour channels.  This results in a total of 256 possible illumination states for each image in the sequence.  Our results show that this reduced search space is sufficient for learning to distinguish very similar objects.

To drive the \gls{SVM} classifier we concatenate features as described in Sect.~\ref{sect_training}, employing a \gls{HOG} feature on a regular grid with cell size 12 and block size 10. We forego the use of a \gls{BoW} as this may be incompatible with classification of visually similar objects, and instead emphasize pose diversity in our training data, and downscale all images to $120 \times 120$ pixels to keep the feature space tractable.

For each illumination matrix evaluated as part of the optimization loop, the classifier training and evaluation are repeated 400 times. The rendered and feature-extracted imagery is randomly split between 75\% training and 25\% test data. Repeating the test makes the most of the dataset, and the average performance is used to drive the optimization process. 

\begin{figure}
    \centering
    \includegraphics[width=\linewidth]{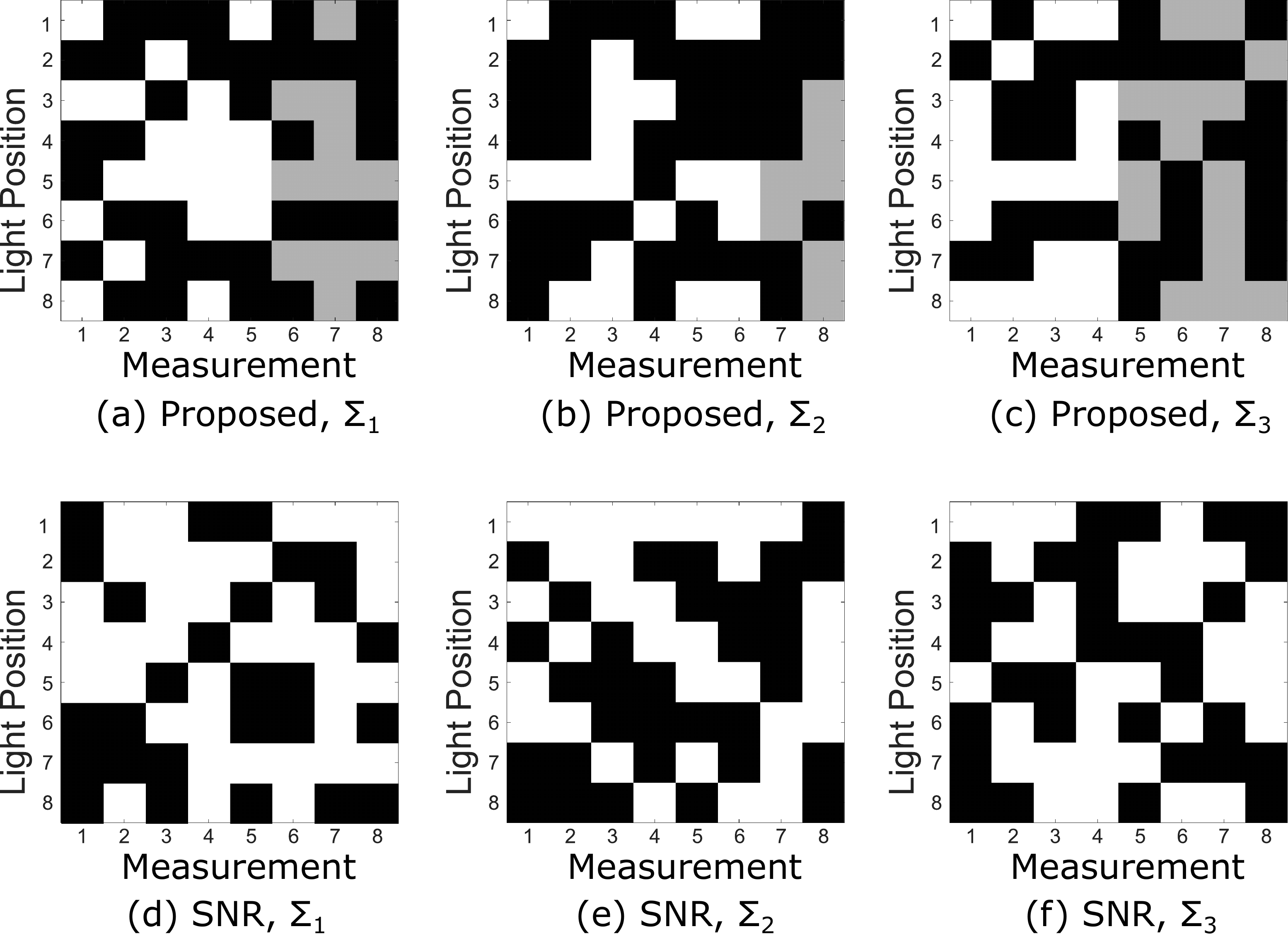}
    \caption{Multiplexing patterns optimized for the three noise characteristics $\Sigma_{1..3}$, with white corresponding to a light being on: (top) patterns optimized with the proposed greedy method, and (bottom) \gls{SNR}-optimal patterns. The proposed method supports early termination of training when performance peaks~-- shaded columns indicate patterns that did not improve performance.}
    \label{fig_result_greedy_mux}
\end{figure}

Carrying out the proposed greedy optimization scheme for each camera setting $\Sigma_{1..3}$ yielded the illumination matrices depicted in Fig.~\ref{fig_result_greedy_mux}. This scheme is capable of early termination, in which training halts once peak performance is reached. However, we completed training across eight patterns for all methods, to provide a more complete evaluation. Patterns that did not result in an increase in performance are grayed out in the figure.  Also shown are the \gls{SNR}-optimal patterns selected for the same camera settings, following the method in Sect.~\ref{sect_snr_optimal_pattern_selection}.

\subsection{Classification Accuracy}

\begin{figure}
    \centering 
    \includegraphics[width=\linewidth]{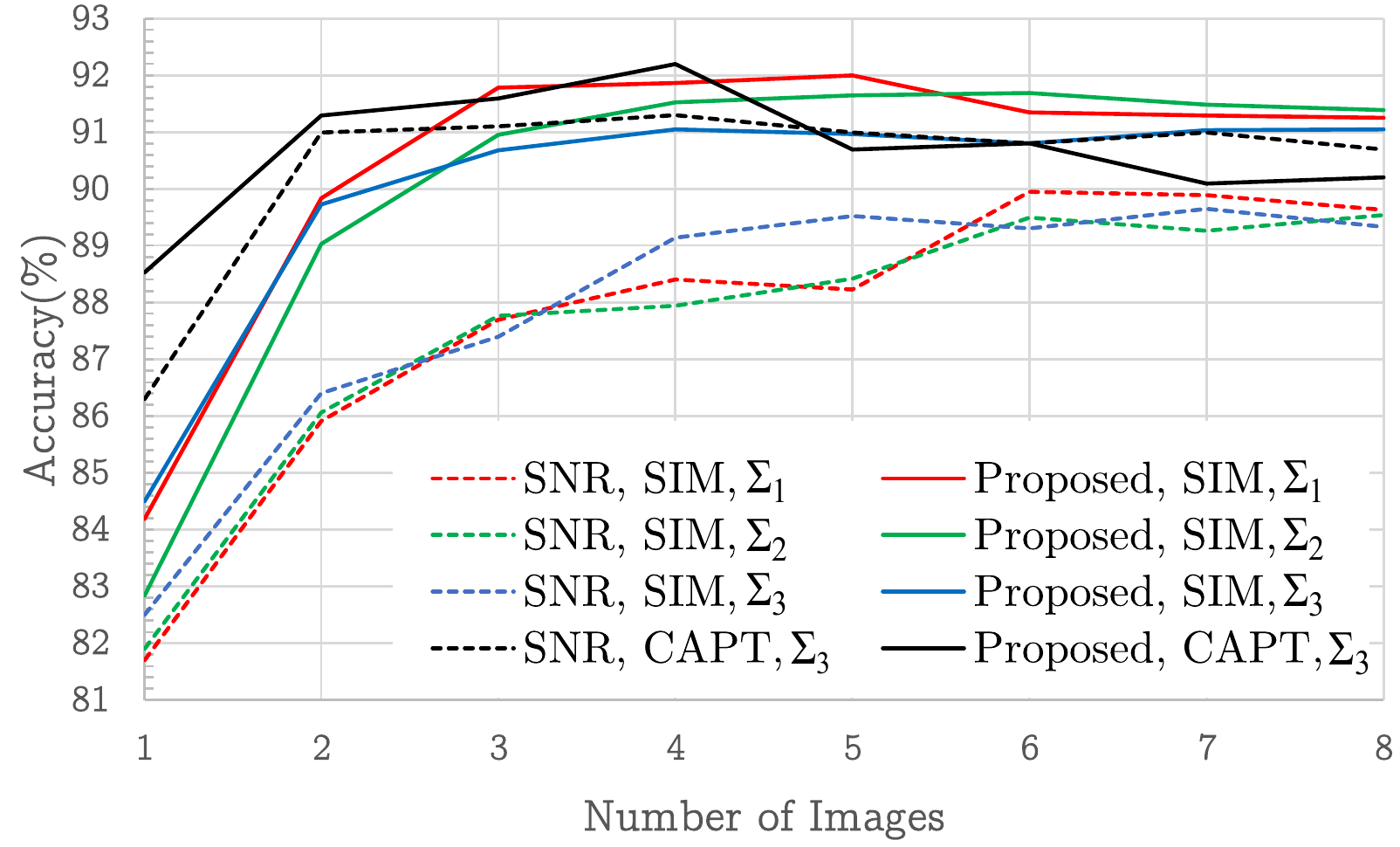}
    \caption{Two experiments on different samples: ``SIM'' uses noise-accurate relit images for both training and testing, while ``CAPT'' is evaluated on previously unseen samples captured using trained illumination patterns.  Accuracy vs.~image count shows the proposed (solid) method outperforming the \gls{SNR}-optimal  (dashed) approach across all three noise characteristics $\Sigma_{1..3}$ with synthetic imagery, and similar performance on captured imagery.  The proposed approach performs better with fewer images, allowing faster performance for a given accuracy.}
    \label{fig_accuracy_vs_imagecount}
\end{figure}

We evaluated the trained illumination patterns shown in Fig.~\ref{fig_result_greedy_mux} classifying synthetic and captured images, with results shown in Fig.~\ref{fig_accuracy_vs_imagecount} and Table~\ref{tab_results}.  Evaluating on simulated imagery involved using the same noise-accurate rendering pipeline used in training to synthesize new images at each of the three camera settings $\Sigma_{1..3}$.  Evaluation on captured imagery involved driving the light stage with the trained illumination patterns to capture previously unseen samples. Noting the similarity between simulation results across settings, we evaluated the captured imagery at only the most challenging setting, $\Sigma_3$.

We evaluated each pattern incrementally, testing first with a single captured image per sample, corresponding to the first column of each pattern, then adding a second image, and so on.  For the proposed method the required classifier for each image count was already available from the pattern selection process. To evaluate the \gls{SNR}-optimal patterns, a classifier was trained for each image count using noise-accurate rendered imagery.

As seen in Fig.~\ref{fig_accuracy_vs_imagecount}, tests on simulated imagery at all three noise levels $\Sigma_{1..3}$ showed the proposed method offering a strong performance advantage over the \gls{SNR}-optimal patterns. The proposed method consistently showed higher accuracy, and grew in accuracy more quickly with image count, reaching peak performance at four or five images, while the \gls{SNR}-optimal method peaked at seven or eight images. Similarity of results across the three noise levels imply the system might be pushed to faster operation. 

For captured imagery, the two methods performed more similarly, with the proposed showing slightly improved performance, especially at low image counts. Importantly, our method was able to perform well despite operating on previously unseen samples captured at the fastest acquisition rate. Both methods performed better than the simulation experiments, and we hypothesize the discrepancy in performance gap is due to the captured-imagery samples being less challenging than the samples evaluated in simulation.

Table~\ref{tab_results} shows a comparison of performance for rendered and captured test imagery at the camera setting $\Sigma_3$. Here we combine synthetic and real samples to report average performance for each of the five types of fruit. Results are shown for the \gls{SNR}-optimal and proposed patterns, as well as for two na\"ive methods.  All methods employed the same \gls{SVM}-based classifier, but the na\"ive approaches did not use multiplexed illumination, instead turning on all LEDs to yield maximal illumination and signal. The na\"ive approach was evaluated both with and without a \gls{BoW}.

The two multiplexing-based methods were evaluated using their optimal image counts.  The proposed method naturally provides for this form of shortcut evaluation, and although the \gls{SNR}-optimal training method does not, the evaluation in Fig.~\ref{fig_accuracy_vs_imagecount} allowed us to select the appropriate image count to give this method the best results. 

The table shows both multiplexing methods significantly outperforming the na\"ive approaches, and the proposed method showing higher overall accuracy than the \gls{SNR}-optimal approach. Training and inference times show the proposed method also allows for faster operation compared with \gls{SNR}-optimal method.  For the \gls{SNR}-optimal approach we show timing results for both the full illumination matrix and the shortcut-evaluated version in parentheses, noting the latter is not obviously attainable using the conventional matrix selection process.  Overall, the proposed method shows the highest accuracy and fastest performance.

\begin{table*}
\small
\caption{Peak performance for the proposed, SNR-optimal, and a na\"ive fixed-illumination approach tested with and without a bag of words (BoW). The na\"ive approaches show weak performance, while the same classifier with no BoW shows much higher performance with both the SNR-optimal and proposed illumination schemes. The proposed method shows the highest overall accuracy, and can reduce training and inference times as peak performance is attained with fewer images. With captured imagery the SNR-optimal method's accuracy peaked at four images, and timing is shown both with and without shortcut evaluation.}
\centering 
\begin{tabular}{ll|llllll|ll}
\Xhline{2\arrayrulewidth}
\multicolumn{1}{c}{\multirow{2}{*}{Approach}} & \multicolumn{1}{c|}{\multirow{2}{*}{Experiment}} & \multicolumn{6}{c|}{Accuracy  (\%)} & \multicolumn{2}{c}{Time (ms)}
\\ 
\multicolumn{1}{c}{} & \multicolumn{1}{c|}{} & Grn Apple & Red Apple & Grape & Banana & Orange & Overall & Train & Infer \\ \Xhline{2\arrayrulewidth}
Na\"ive & SIM  & 43.75 & 67.50 & 72.50 & 56.25 & 67.50 & 61.50 & - & - \\ \hline
Na\"ive BoW & SIM  & 51.25 & 52.50 & 93.75 & 63.75 & 68.75 & 66.00 & - & - \\ \hline
SNR & SIM  & 89.25 & 91.70 & \textbf{96.65} & 71.90 & \textbf{99.55} & 89.81 & 1197 & 35 \\ \hline
Proposed & SIM  & \textbf{91.65} & \textbf{93.50} & 96.15 & \textbf{74.50} & 99.45 & \textbf{91.05} & \textbf{650} & \textbf{24} \\ \Xhline{2\arrayrulewidth}
SNR & CAPT & 84.15 & \textbf{95.00} & \textbf{98.50} & 79.75 & \textbf{99.25} & 91.33 & 1276 (654)  & 756 (376)\\ \hline
Proposed & CAPT & \textbf{94.20} & 90.70 & 96.85 & \textbf{83.45} & 95.80 & \textbf{92.20} & 651 & 376\\ \Xhline{2\arrayrulewidth}
\end{tabular}
\label{tab_results}
\end{table*}

\section{Conclusions and Future Work}
\label{sect_conclusions}

In this paper we presented an active illumination-based means of making a very challenging visual classification task achievable with even a conventional \gls{SVM}-based classifier.  We applied light stage technology in two ways: driving inference-time illumination patterns to allow fast classification of visually similar objects, and accurately rendering relit samples to drive pattern selection and classifier training.  

We constructed a compact RGB-IR light stage and used it to classify visually similar objects.  We compared against na\"ive approaches, establishing that without the light stage the classification task is not well achieved. We then proposed two alternative illumination pattern selection schemes, one that optimizes \gls{SNR} in the conventional multiplexing sense \citep{schechner2007multiplexing}, and one greedy optimization scheme that incrementally grows a pattern while training a matching classifier.  We showed the greedy approach achieves better overall performance on simulated and captured test samples, and performs better with fewer images than the \gls{SNR}-optimal pattern.

Although specialized illumination hardware is required to carry out our proposed training and classification scheme, this might find application in a range of situations where conventional approaches cannot effectively operate, and that require fast classification of visually similar samples. Examples are in forgery detection, disease detection for crops and plants, diagnosis of dermal lesions and spots, and quality control for manufacturing and agriculture. 

We see several avenues for improvement and hope this work will inspire further development. The pattern selection process only considered the eight illumination positions, repeating the pattern over colour channels. Jointly training chromatic and spatial domains yields a larger search space but should also offer faster, more accurate results.  Extension to consider more sophisticated classifiers is also possible, and we expect that jointly learning illumination patterns, feature extraction and classification might allow operation on even more challenging cases. 
{\small
\bibliographystyle{model2-names}
\bibliography{refs}}
\end{document}